\documentclass[journal]{IEEEtran}

\usepackage{cite}

\ifCLASSINFOpdf
  \usepackage{graphicx}
\else
\fi
\usepackage{amsmath}
\usepackage{amssymb}

 \usepackage[caption=false,font=footnotesize]{subfig}
\usepackage{url}

\usepackage{multirow}
\usepackage{booktabs}
\usepackage{xcolor}

\begin{document}
%

\title{Corrupting Attention: Evasion-Based Adversarial Attacks on Encoder Attention in Detection Transformers}

\author{
    Ridma Jayasundara\textsuperscript{\rm 1}, 
    Shaheer Mohamed\textsuperscript{\rm 1},
    Tharindu Fernando\textsuperscript{\rm 1},
    Harshala Gammulle\textsuperscript{\rm 1},
    Basura Fernando\textsuperscript{\rm 2},
    Sanka Rasnayake\textsuperscript{\rm 3},
    A V Subramanyam\textsuperscript{\rm 4},
    Sridha Sridharan\textsuperscript{\rm 1},
    Clinton Fookes\textsuperscript{\rm 1}%
\thanks{\textsuperscript{\rm 1}Signal Processing, Artificial Intelligence and Vision Technologies Group (SAIVT), Queensland University of Technology}%
\thanks{\textsuperscript{\rm 2}Institute of High-Performance Computing, Agency for Science, Technology and Research, Singapore}%
\thanks{\textsuperscript{\rm 3}School of Computing, National University of Singapore}%
\thanks{\textsuperscript{\rm 4}Indraprastha Institute of Information Technology, Delhi, India}%
}

\maketitle

\begin{abstract}
Adversarial vulnerabilities remain a major concern for the safe deployment of neural networks, particularly in object detection, a core task embedded in many safety-critical systems. Detection transformers have emerged as leading object detectors, yet their adversarial robustness remains comparatively underexplored. Most existing attacks target the detection output rather than the attention mechanism that makes these models distinctive. In this paper, we introduce the first attack that directly optimizes an encoder-attention objective under an imperceptible, bounded $\ell_\infty$ perturbation. Rather than introducing an attacker-owned sink token through a visible patch, it drives the model's own attention toward a corrupted target. We argue that encoder attention concentrates the model's spatial reasoning, so corrupting it propagates through the detection pipeline more disruptively than perturbing the detection output alone. Our attack reduces DETR-R50 mAP on COCO from 42.1 to 0.97, a $\sim 4\times$ reduction in resulting mAP over the strongest existing attack under an identical perturbation budget and iteration count. We further show that this vulnerability is not specific to a particular corruption objective: across four qualitatively distinct targets, dispersion, re-ranking, permutation, and peak-suppression, detection consistently drops below 3 mAP, suggesting that the weakness arises from disrupting the attention structure itself rather than from any single target. Finally, we demonstrate that the attack generalizes across attention formulations, reducing DINO-Swin-L from 56.8 to 1.44 mAP against 7.3 for the strongest prior attack, establishing state-of-the-art on both dense and deformable attention.
\end{abstract}


%
\IEEEpeerreviewmaketitle

\section{Introduction}
\IEEEPARstart{O}{bject} detection is one of the most fundamental and consequential tasks in computer vision, as it jointly answers two complementary questions: what objects are present in an image and where they are located~\cite{zaidi2022survey}. It is also a main component of many real-world and safety-critical systems, which makes its adversarial robustness a practical concern. Modern deep learning detectors are commonly organised into three broad families: two-stage detectors, single-stage detectors, and, more recently, transformer-based detectors~\cite{zaidi2022survey,arkin2023survey}. The first two are built on convolutional neural network (CNN) backbones and dominated the field for the better part of a decade. The third was pioneered by the Detection Transformer (DETR)~\cite{carion2020end}, which recast detection as a direct set prediction problem and replaced components such as anchor generation and non-maximum suppression with an encoder-decoder architecture. With transformers becoming increasingly dominant, subsequent variants of this architecture~\cite{zhu2020deformable,zhang2022dino} now define the state of the art on major detection benchmarks. Yet their adversarial robustness remains underexplored, as most attack research
has focused on CNN-based detectors~\cite{nazeri2026evaluating,yahn2025adversarial}, and these attacks transfer poorly across the architectural gap.

\begin{figure}[t]
\centering
\includegraphics[width=\columnwidth]{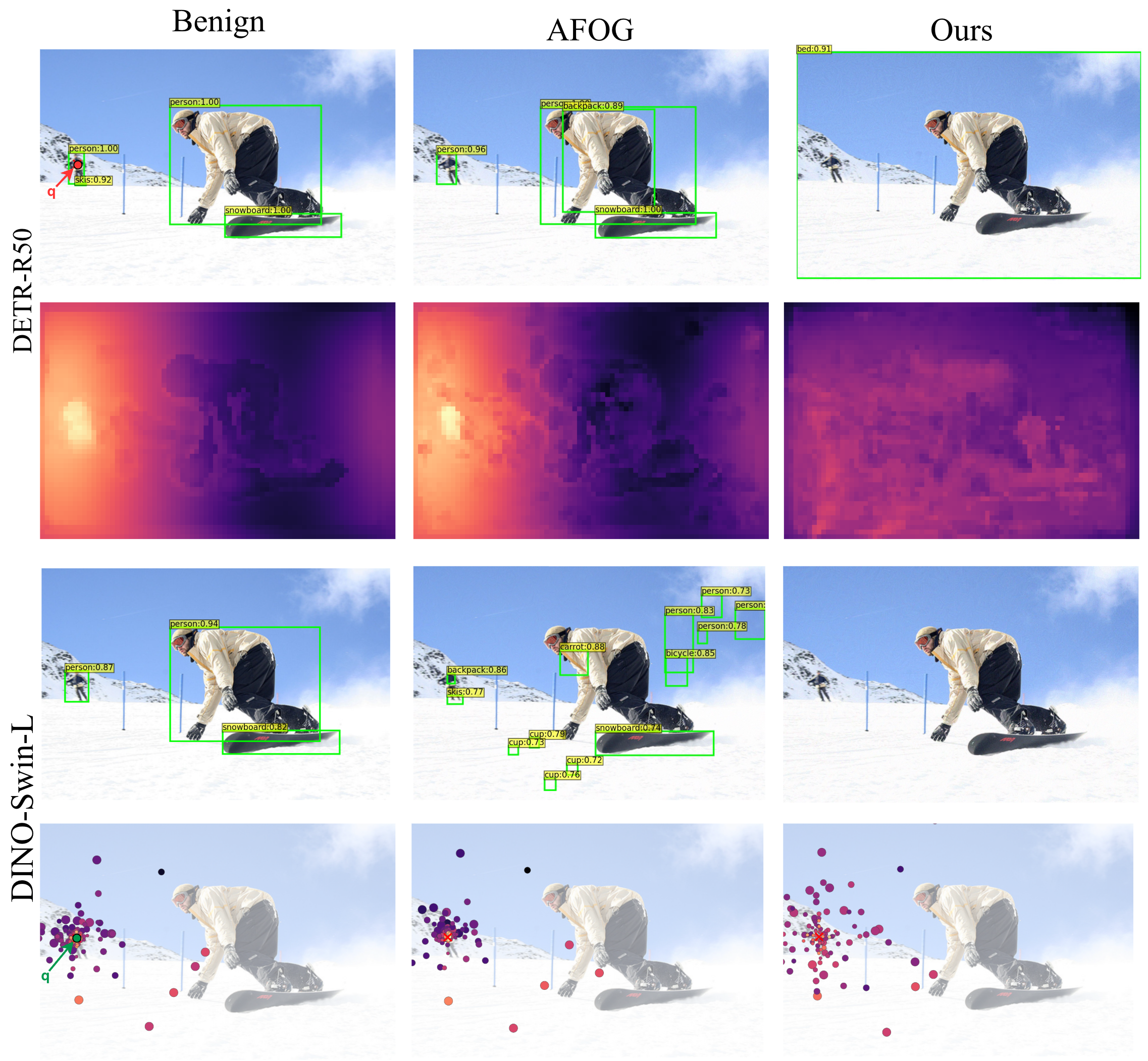}
\caption{Detections and final layer encoder attention visualization for DETR-R50 (top two rows) and DINO-Swin-L (bottom two rows) on a benign image, under AFOG~\cite{yahn2025adversarial}, and our attack. The direct corruption of attention in our attack collapses the detection more effectively.}
\label{fig:hero}
\end{figure}

Across the DETR family, the encoder occupies a common and pivotal position. Regardless of the backbone or decoder query design, current state-of-the-art DETR variants pass backbone features through a stack of attention-based encoder layers before producing any detections~\cite{li2023lite}. The encoder refines the backbone features into context-aware representations, modelling the relationships between spatial locations that the decoder subsequently queries to localise and classify objects~\cite{li2023lite}.

The mechanism used to compute these relationships is precisely what distinguishes the variants. The original DETR employs global self-attention in which every spatial location attends to every other~\cite{carion2020end}. In contrast, Deformable DETR and its descendants, including DINO~\cite{zhang2022dino}, replace this with deformable attention~\cite{zhu2020deformable}, restricting each query to a small, learnable set of sampling points and thereby reducing the quadratic attention cost to linear. Whichever mechanism is used, the encoder refines the backbone features into the memory that every decoder query reads. If this memory is corrupted, the decoder inherits degraded representations no matter how capable the backbone or decoder, and detection collapses. 

Existing adversarial attacks do not target this vulnerability. Most were designed for convolutional detectors and transfer poorly to transformer-based ones. Those built for transformers optimize against the detection output, leaving the attention mechanism unexploited~\cite{yahn2025adversarial,nazeri2026evaluating}. A separate line of work does target attention directly, but does so through a visible adversarial patch that introduces an external, attacker-owned sink token~\cite{lovisotto2022give,fu2022patch}, rather than an imperceptible perturbation of the image. These attacks are also confined to global dot-product attention and do not carry over to the deformable attention of modern variants.

In this paper, we introduce the first attack on object detectors that optimizes an encoder-attention objective directly under an imperceptible, bounded
$\ell_\infty$ perturbation. Rather than introducing an attacker-owned sink token, we drive the model's own attention toward a structured corruption target. We define four such targets: dispersion, re-ranking, permutation, and peak-suppression, each destroying a different structural property of the attention row, and show empirically that detection collapses under all of them, losing over $93\%$ of clean mAP on DETR-R50. This indicates the vulnerability lies in the attention itself rather than in any particular choice of corruption target. Because we target the attention map itself and not the mechanism that produces it, the attack extends to different attention formulations. We show this on DINO's deformable attention, where it reduces DINO-Swin-L from $56.8$ to $1.44$ mAP. Figure~\ref{fig:hero} illustrates this. Under an identical perturbation budget, the strongest prior attack leaves the encoder's attention partly intact and several objects still detected, whereas driving the attention at object tokens toward a corrupted target collapses both the attention and the detections, for global and deformable attention alike. The perturbations further transfer between the two architectures (DETR-R50 $\leftrightarrow$ DINO-R50), again outperforming prior work.

The main contributions of this work are as follows.
\begin{itemize}
    \item We propose a direct attention corruption attack on transformer-based detectors, optimizing an input-space perturbation against the encoder's attention rather than the detection loss. Under an identical budget and iteration count, it reduces DETR-R50 from $42.1$ to $0.97$ mAP and DINO-Swin-L from $56.8$ to $1.44$, establishing a new state of the art on both.
    
    \item We show the vulnerability is robust to the choice of corruption, across four qualitatively distinct objectives, namely dispersion, re-ranking, permutation, and peak-suppression. All four collapse detection and achieve SOTA on DETR-R50, and three of them do so on DINO as well, showing that the weakness lies in the attention structure itself.

    \item We demonstrate that the attack generalizes across different attention formulations, extending from DETR's dense self-attention to DINO's deformable attention, indicating the attack surface is not tied to a specific attention implementation.
\end{itemize}

\section{Related Work}

\subsection{Adversarial Attacks on Object Detectors}
Adversarial attacks on object detection were first developed for convolutional detectors, where perturbations target region proposals, anchors, or objectness scores. DAG~\cite{xie2017adversarial}, UEA~\cite{wei2018transferable}, RAP~\cite{li2018robust}, and TOG~\cite{chow2020adversarial} all follow this recipe, degrading Faster R-CNN, SSD, and YOLO. Transformer-based detectors expose no such surface. Predictions arise from one-to-one set matching over a sparse set of queries, with no dense objectness map and no Non-Maximum Suppression. Attacks built on CNN architecture are therefore substantially less effective in the transformer setting~\cite{yahn2025adversarial}.

Pixel-level gradient attacks such as FGSM~\cite{goodfellow2014explaining}, PGD~\cite{madry2018towards} and C\&W~\cite{carlini2017towards} have been adapted to DETR~\cite{carion2020end}. More recently, and Nazeri et al.~\cite{nazeri2026evaluating} propose a DETR-specific C\&W variant that exploits intermediate decoder losses. Their analysis further finds that DETR variants are broadly vulnerable across attack types, with adversarial examples transferring readily among DETR variants but far less effectively to convolutional detectors such as Faster R-CNN, highlighting the architecture-specific nature of these perturbations. AFOG~\cite{yahn2025adversarial} introduces an imperceptible, bounded perturbation attack tailored to detection transformers, using a learnable input space weighting map to concentrate the perturbation on vulnerable image regions. Notably, it reports that the encoder's self-attention collapses under attack, yet this collapse is only a byproduct of optimizing a detection loss. The attention mechanism is never targeted directly. These works confirm that detection transformers are vulnerable to bounded perturbations, but they treat the detector as a generic loss surface and leave the attention mechanism untouched, which is the component that distinguishes these architectures from ours.

\subsection{Attacking the Attention Mechanism}
A second line of work targets attention directly, using adversarial patches. Attention-Fool~\cite{lovisotto2022give} tries to optimize a patch in a way that nearly all queries attend to the single token covered by the patch, diverting attention away from the real objects. It was originally developed for Vision Transformer (ViT) \cite{dosovitskiy2020image} and Data-efficient Image Transformer (DeiT) \cite{touvron2021training} for image classification and was later adapted to DETR. Patch-Fool~\cite{fu2022patch} similarly optimizes a patch to maximize the attention it receives from all other tokens, but is evaluated only on image classification rather than object detection. LQA~\cite{wang2026localized} carries the same idea into detection transformers, hijacking both encoder self-attention and decoder cross-attention through an on-object patch. The common thread is that all of these methods redirect attention toward an external sink token introduced by a visible, localized patch.

Such attacks are specific to global dot product attention. Deformable attention, introduced to reduce the quadratic cost of global attention in detectors such as Deformable-DETR and DINO, breaks their central assumption. In deformable attention, each query attends to only a few dynamically sampled keys and no single key is shared across queries, so one hijacked token can no longer capture all queries. To alleviate this issue, Alam et al.~\cite{alam2025adversarial} extend the patch-based recipe specifically for deformable attention by fabricating a sink, using a source patch that redirects the sampling offsets of nearby tokens toward a second target patch carrying the adversarial noise.

Across this literature, each prior attack on attention exhibits at least one of three limitations: it relies on a visible, localised patch that introduces an attacker owned sink token; it applies only to global dot product attention and does not carry over to the deformable attention of modern detectors; or, as with the detection loss perturbations above, it corrupts attention only as an incidental effect rather than optimising it directly. In contrast, we introduce the first attack that directly optimises an encoder attention objective under an imperceptible, bounded $\ell_\infty$ perturbation, with no adversarial patch and no sink token. It drives the model's own attention toward a corrupted target of its clean distribution, and because the objective acts only on the attention weights, it is well defined for both global and deformable attention.

\section{Methodology}

Multi-head self-attention is the main component of the detection transformer encoder, and its sensitivity to input is theoretically established in~\cite{kim2021lipschitz}, which proves that standard dot-product self-attention is not globally Lipschitz, indicating that it can be highly sensitive to its input representations. While this result concerns attention's sensitivity to its input features and bounds a worst-case rate rather than any specific perturbation, it motivates our main idea. Instead of attacking the detector's output, as most existing attacks do, we target the encoder self-attention directly, and show empirically that an imperceptible image-space perturbation corrupts it and degrades detection. 

\begin{figure}[t]
    \centering
    \includegraphics[width=\columnwidth]{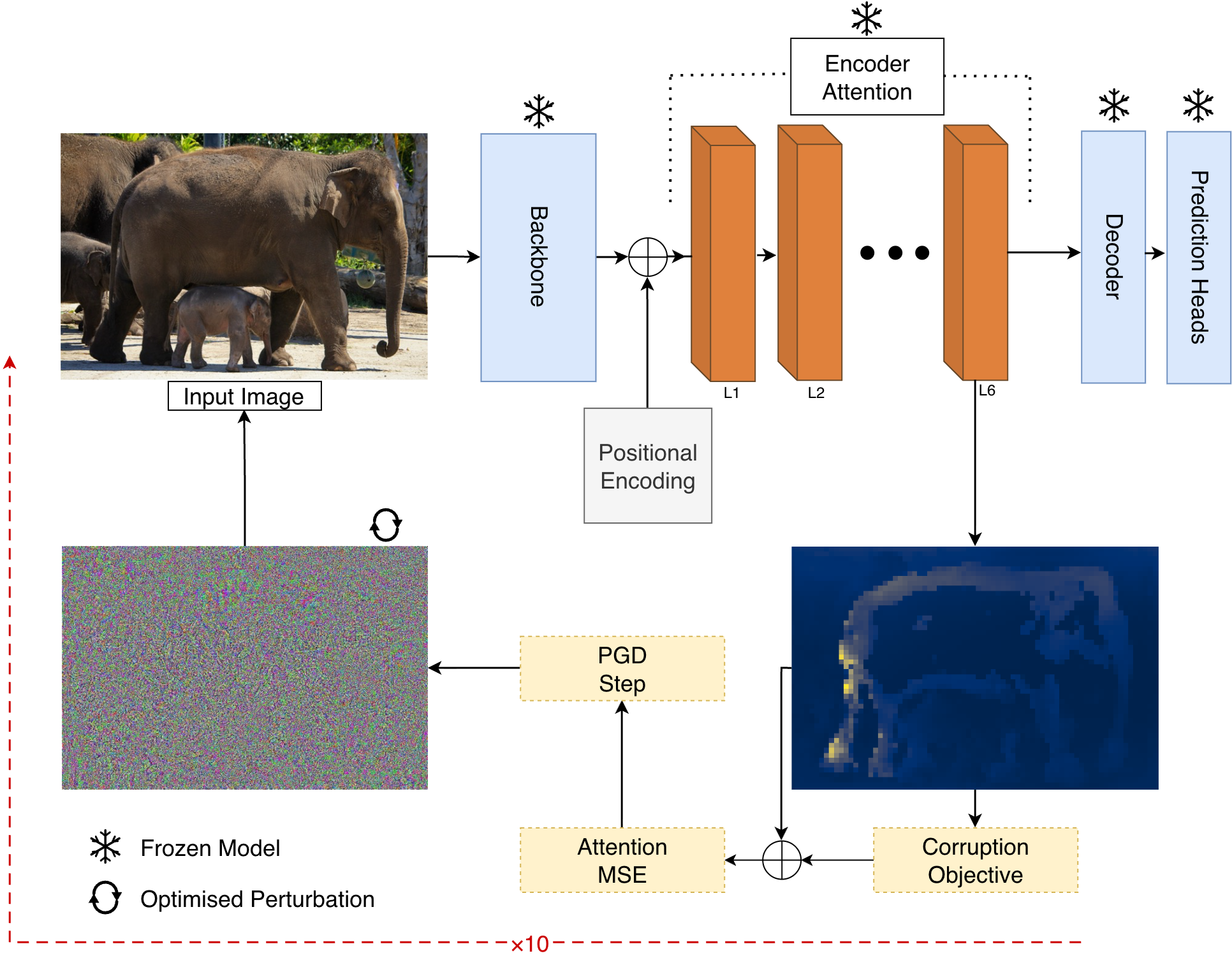}
    \caption{Attack framework. The corruption objective turns the clean final-layer encoder attention into a fixed target, and PGD optimises an $\ell_\infty$ bounded perturbation against it. The model is frozen throughout.}
    \label{fig:optimization_pipeline}
\end{figure}

We construct an input-space perturbation that drives the encoder's native attention distribution toward a structured corruption target, and find that several structurally distinct targets substantially degrade detection, indicating that the vulnerability lies in disrupting attention. The overall attack framework is illustrated in Figure~\ref{fig:optimization_pipeline}. In the remainder of this section, we explain our method in detail.

\subsection{Problem Formulation}

Let $f_\theta$ denote the victim detector with frozen parameters $\theta$. It maps an
input image to a set of detections through three stages, where $f_\theta = \mathrm{Dec}_\theta \circ \mathrm{Enc}_\theta \circ g_\theta$ as shown in Eq.~\ref{eq:pipeline},

\begin{equation}
    x \;\xrightarrow{\;\; g_\theta \;\;}\; X \in \mathbb{R}^{S \times d}
    \;\xrightarrow{\;\; \mathrm{Enc}_\theta \;\;}\; \tilde{X}
    \;\xrightarrow{\;\; \mathrm{Dec}_\theta \;\;}\; \hat{y}
    \label{eq:pipeline}
\end{equation}
Here, $x \in \mathbb{R}^{3 \times H \times W}$ is the input image. The backbone $g_\theta$ maps it to a feature grid of size $h \times w$, which is flattened into a sequence of $S = hw$ feature tokens, each of dimension $d$. After adding positional embeddings, the resulting token sequence $X \in \mathbb{R}^{S \times d}$ is passed to the encoder. The encoder $\mathrm{Enc}_\theta$ refines these tokens through $L$ layers of transformer blocks with multi-head self-attention, producing $\tilde{X} \in \mathbb{R}^{S \times d}$; and the decoder $\mathrm{Dec}_\theta$ produces the final detections $\hat{y}$ from a set of learned queries. Our perturbation acts on $x$, at the input to $g_\theta$, while our objective is defined on the attention inside $\mathrm{Enc}_\theta$.

Within an encoder layer, multi-head self-attention~\cite{vaswani2017attention} is computed for each head $m \in \{1,\dots,n_h\}$ as shown in Eq.~\ref{eq:attention},
\begin{equation}
    Z^{(m)} = \frac{Q^{(m)} \big(K^{(m)}\big)^{\top}}{\sqrt{d_h}},
    \qquad
    A^{(m)} = \mathrm{softmax}\big(Z^{(m)}\big),
    \label{eq:attention}
\end{equation}

with $Q^{(m)} = X W_Q^{(m)}$, $K^{(m)} = X W_K^{(m)}$, $V^{(m)} = X W_V^{(m)}$, head dimension $d_h = d / n_h$, and the softmax applied row-wise. We refer to $Z^{(m)}$ as the attention logits and $A^{(m)}$ as the attention map. Each row $A^{(m)}_{t:}$ is a probability distribution over the $S$ tokens, and the output of each head is $O^{(m)} = A^{(m)} V^{(m)}$, so that the output at token $t$ is
$\mathbf{o}^{(m)}_t = \sum_{s} A^{(m)}_{ts}\,\mathbf{v}^{(m)}_s$. This distribution determines the spatial context that position aggregates. We define our attack on the logits $Z^{(m)}$ rather than the map $A^{(m)}$, since gradients reaching the input through post-softmax weights are strongly attenuated, making $A^{(m)}$ a weak optimization target~\cite{lovisotto2022give}, which we also confirm empirically.

Rather than corrupting every row, we target only the query positions corresponding to detected content. We run a clean forward pass, keep the detections scoring above a threshold $\tau$, and take as our target set $\mathcal{O} = \{\, t : c_t \in \textstyle\bigcup_i B_i \,\}$ the tokens $t$ whose spatial location $c_t$ falls inside a predicted box $B_i$. 

Let $Z^{(m)}_{t:}(x)$ denote the logit row at object token $t$ and head $m$ at a chosen
encoder layer. A corruption operator $T$ maps a clean row to a corrupted target,
$Z^{\star(m)}_{t:} = T\big(Z^{(m)}_{t:}(x)\big)$, computed once from the clean image and
held fixed during the attack. We seek an additive perturbation $\delta$, bounded in
$\ell_\infty$ by $\epsilon$, that drives the attention at the object tokens toward these targets, as shown in Eq.~\ref{eq:objective},
\begin{equation}
    \delta^{\star} =
    \operatorname*{arg\,min}_{\|\delta\|_\infty \le \epsilon}
    \underbrace{
      \frac{1}{n_h |\mathcal{O}|}
      \sum_{m=1}^{n_h} \sum_{t \in \mathcal{O}}
      \mathcal{L}\!\left( Z^{(m)}_{t:}(x + \delta), \; Z^{\star(m)}_{t:} \right)
    }_{\mathcal{L}_{\mathrm{total}}(x+\delta;\, Z^{\star})}.
    \label{eq:objective}
\end{equation}
Notably, the objective contains no detection loss and the attack is blind to the detector's output during optimization. We define the family of operators $T$ next.

\begin{figure}[!t]
\centering
\subfloat[Corruption of the object rows.]{%
  \includegraphics[width=0.45\textwidth]{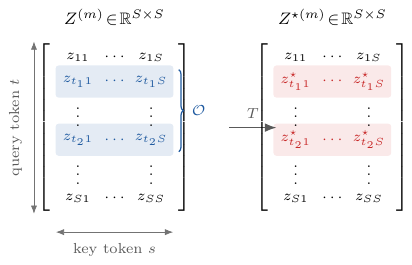}%
  \label{fig:matrices}}
\\
\subfloat[The four corruption targets.]{%
  \includegraphics[width=0.45\textwidth]{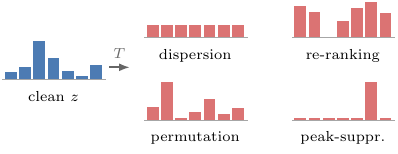}%
  \label{fig:targets}}
\caption{The attention corruption attack. (a) The operator $T$ replaces the attention
logit rows of the object tokens $\mathcal{O}$, leaving every other row and all value
vectors unchanged. (b) A clean row and the four targets, each destroying a different
structural property of the distribution: concentration, ordering, key assignment, and
the dominant mode.}
\label{fig:attack}
\end{figure}
\subsection{Corruption Targets}
\label{sec:targets}

A corruption operator $T$ takes a clean logit row and returns a target row that no longer routes the query toward the keys it should attend to. To test whether this vulnerability depends on how the attention is disrupted, we design four qualitatively distinct operators as shown in Figure~\ref{fig:attack}, each corrupting the row in a different way. As we show, all four collapse detection, showing that the weakness lies in attention itself. Throughout, we write $z = Z^{(m)}_{t:}$ for a clean logit row and $z^{\star}$ for its corrupted target.

\subsubsection{Dispersion}
The most direct corruption is to remove all concentration from the row, spreading attention uniformly across every key. Dispersion sets every logit equal,

\begin{equation}
    z^{\star}_{s} = 0 \quad \forall s \in \{1,\dots,S\},
    \label{eq:dispersion}
\end{equation}
so that $A^{\star}_{s} = 1/S$ and
$\mathbf{o}^{(m)}_t = \bar{\mathbf{v}}^{(m)}$ for every targeted token
$t \in \mathcal{O}$. The attention branch then supplies the same pooled context
to all object tokens in place of their position-specific context.

\subsubsection{Re-ranking}
Re-ranking reverses the order of the logits, turning the most-attended key into the least-attended and vice versa,
\begin{equation}
    z^{\star}_{s} = \max_{s' \in \{1,\dots,S\}} z_{s'} - z_{s},
    \label{eq:reranking}
\end{equation}
so the keys a query previously attended become the ones it attends to least.

\subsubsection{Permutation}
Permutation shuffles the logits by a random permutation $\pi$ of the key positions,
\begin{equation}
    z^{\star}_{s} = z_{\pi(s)},
    \label{eq:permutation}
\end{equation}
so the distribution keeps its shape but each weight now lands on an unrelated key. Although self-attention is permutation-equivariant, permuting only the logits while holding the value vectors fixed sends each query's weights to mismatched content.

\subsubsection{Peak-suppression}
Peak-suppression moves the row's dominant peak onto the least-attended key and flattens everything else to the row minimum,
\begin{equation}
   z^{\star}_{s} =
\begin{cases}
    \max_{s' \in \{1,\dots,S\}} z_{s'} & s = \arg\min_{s' \in \{1,\dots,S\}} z_{s'}, \\[2pt]
    \min_{s' \in \{1,\dots,S\}} z_{s'} & \text{otherwise}.
\end{cases}
    \label{eq:peak-suppression}
\end{equation}
so the query attends most strongly to the key it should attend to least.

\subsection{Perturbation Optimization}
\label{sec:optimization}
We solve the objective in Eq.~\ref{eq:objective} with projected gradient descent under an $\ell_\infty$ constraint~\cite{madry2018towards}. We instantiate the Loss $\mathcal{L}$ as the mean squared error (MSE) between the perturbed and target logit rows as in Eq.~\ref{eq:loss},

\begin{equation}
    \mathcal{L}\big(Z^{(m)}_{t:}(x+\delta),\, Z^{\star(m)}_{t:}\big)
    = \frac{1}{S}\sum_{s}
      \Big( Z^{(m)}_{ts}(x+\delta) - Z^{\star(m)}_{ts} \Big)^{2},
    \label{eq:loss}
\end{equation}

Here we evaluate the loss per head and per object token, so that every head's attention is corrupted rather than only an averaged map. The targets $Z^{\star(m)}_{t:}$ are constructed once from a clean forward pass and held fixed throughout the iterations.

The perturbation is initialized uniformly within the budget,
$\delta_0 \sim \mathcal{U}(-\epsilon,\epsilon)$, and updated at each iteration $k$ by a
signed gradient step followed by projection onto the $\epsilon$-ball,
\begin{equation}
    \delta_{k+1} \gets \Pi_{\epsilon}\!\left[\,
      \delta_{k} - \alpha\,\Gamma\!\left(
      \frac{\partial \mathcal{L}_{\mathrm{total}}(x + \delta_k;\, Z^{\star})}
           {\partial \delta_k}\right)\right],
    \label{eq:pgd}
\end{equation}
where $\alpha$ is the perturbation step size, $\Gamma$ is the sign function, and
$\Pi_{\epsilon}$ clips each coordinate of $\delta$ to $[-\epsilon,\epsilon]$. The
perturbed image is clipped to the valid pixel range after each step. Model parameters $\theta$ remain frozen; gradients are taken only with respect to the input and propagate from the encoder attention logits back through the backbone.

\subsection{Extension to Deformable Attention}
\label{sec:deformable}

Deformable attention~\cite{zhu2020deformable}, used by DINO~\cite{zhang2022dino},
replaces the dense query-key interaction with a sparse one. For a query token $q$ with reference point $p_q$, a linear projection of the query feature predicts $K$ sampling offsets $\Delta
p_{qk}$ together with their weights, which are normalized by a softmax over the $K$
points. The output is
\begin{equation}
    \mathrm{DeformAttn}(z_q, p_q, x)
    = \sum_{k=1}^{K} A_{qk}\; x\big(p_q + \Delta p_{qk}\big),
    \label{eq:deform}
\end{equation}

Here, $\sum_{k=1}^{K} A_{qk} = 1$. The weights are therefore produced by a projection of the query rather than by an
affinity computation between tokens, but their role is the same as in Eq.~\ref{eq:attention}. $A_{q:}$ is a normalized distribution determining how a query aggregates content from the positions it reads. The difference is that it reads from
$K \ll S$ learned locations rather than from every token. Since our operators act on this distribution, they apply directly with $S$ replaced by $K$, and the objective and optimization procedure carry over unmodified. Because the attack targets the weight distribution rather than the computation that produces it, it transfers to the deformable setting unchanged.

\section{Experiments}
\label{sec:exp}

\subsection{Experimental Setup}
\label{sec:setup}

All experiments are conducted on the validation split of MS COCO 2017~\cite{lin2014microsoft}, which has $5{,}000$ images annotated over $80$ object categories.  Detection quality is measured with mean average precision (mAP) averaged over IoU thresholds from $0.50$ to $0.95$ in steps of $0.05$, the primary COCO metric. As ours is an evasion attack, a lower adversarial mAP indicates a stronger attack. All models and attacks are implemented in PyTorch~\cite{paszke2019pytorch} and all experiments run on a NVIDIA RTX A6000 GPU. We follow the same experimental setup as~\cite{yahn2025adversarial}. The perturbation is optimized under a budget of $\epsilon = 8/255$ with a step size of $2/255$ for $10$ iterations, and object regions are selected using a confidence threshold of $\tau = 0.5$.

\subsection{Comparison with State-of-the-Art Attacks}
\label{sec:sota}

We compare our attack with different state-of-the-art attacks on DETR-R50 and DINO-Swin-L and the results are shown in Table~\ref{tab:attack_comparison}. The baseline results are grouped by how gradients are obtained. Surrogate attacks build the perturbation on a substitute model and transfer it to the victim, whereas victim attacks, including ours, explicitly build the attack on the victim model itself. We list each attack's perturbation budget and iteration count along with its adversarial mAP, for a fair comparison.

\begin{table}[t]
\caption{Comparison against state-of-the-art detection attacks on DETR-R50 and DINO-Swin-L. Sur./Vic.\ denote surrogate and victim attacks. Ours uses dispersion on DETR and re-ranking on DINO. Baseline numbers as reported in~\cite{yahn2025adversarial}.}
\label{tab:attack_comparison}
\centering
\small
\setlength{\tabcolsep}{3pt}
\begin{tabular}{lccccc}
\toprule
\multirow{2}{*}{Attack} &
\multirow{2}{*}{Type} &
\multirow{2}{*}{Pert. Budget} &
\multirow{2}{*}{Iters.} &
\multicolumn{2}{c}{Adversarial mAP} \\
\cmidrule(lr){5-6}
& & & & DETR & DINO \\
\midrule
GARSDC   & Sur. & 0.05  & 3000+   & 6.0  & --   \\
GALD     & Sur. & 0.063 & 10      & 20.6 & --   \\
RAD      & Sur. & 0.063 & 10      & 27.2 & 47.2 \\
GHFD     & Sur. & 0.063 & 50      & 12.7 & 42.3 \\
UEA      & Sur. & 0.063 & 50      & 28.5 & 50.7 \\
DAG      & Sur. & 0.063 & 50      & 28.6 & 50.7 \\
RAP      & Sur. & 0.063 & 50      & 24.7 & 49.5 \\
EBAD     & Vic. & 0.039 & 10      & 34.9 & --   \\
AttentionFool & Vic. & --    & 10--150 & 21.0 & --   \\
OATB     & Vic. & 0.078 & 20      & 26.6 & --   \\
AFOG     & Vic. & 0.031 & 10      & 4.1  & 7.3  \\
\midrule
\textbf{Ours} & \textbf{Vic.} & \textbf{0.031} & \textbf{10} &
\textbf{0.97} & \textbf{1.44} \\
\bottomrule
\end{tabular}
\end{table}

\begin{table}[t]
\caption{Effectiveness of AFOG and our attack across eight detection transformers, measured by mAP on perturbed images. Dispersion is used as the corruption target throughout.}
\label{tab:transformer_backbones}
\centering
\small
\setlength{\tabcolsep}{4pt}
\begin{tabular}{lcccc}
\toprule
Model & Params (M) & Benign & AFOG & \textbf{Ours} \\
\midrule
DETR-R50*   & 39.8   & 42.1 & 4.1          & \textbf{0.97}  \\
DETR-R101*  & 76.0   & 43.5 & 5.2          & \textbf{0.93}  \\
R50         & 47.6   & 49.2 & 5.3          & \textbf{4.18}  \\
ConvNeXt    & 219.0  & 55.4 & \textbf{3.9} & 3.98           \\
Swin-L      & 217.2  & 56.8 & 7.3          & \textbf{4.73}  \\
InternImage & 241.0  & 56.9 & 7.3          & \textbf{5.75}  \\
FocalNet    & 228.9  & 58.5 & 7.3          & \textbf{4.12}  \\
EVA         & 1037.2 & 62.1 & 12.2         & \textbf{11.56} \\
\bottomrule
\end{tabular}
\end{table}

\begin{figure*}[t]
\centering
\includegraphics[width=\textwidth]{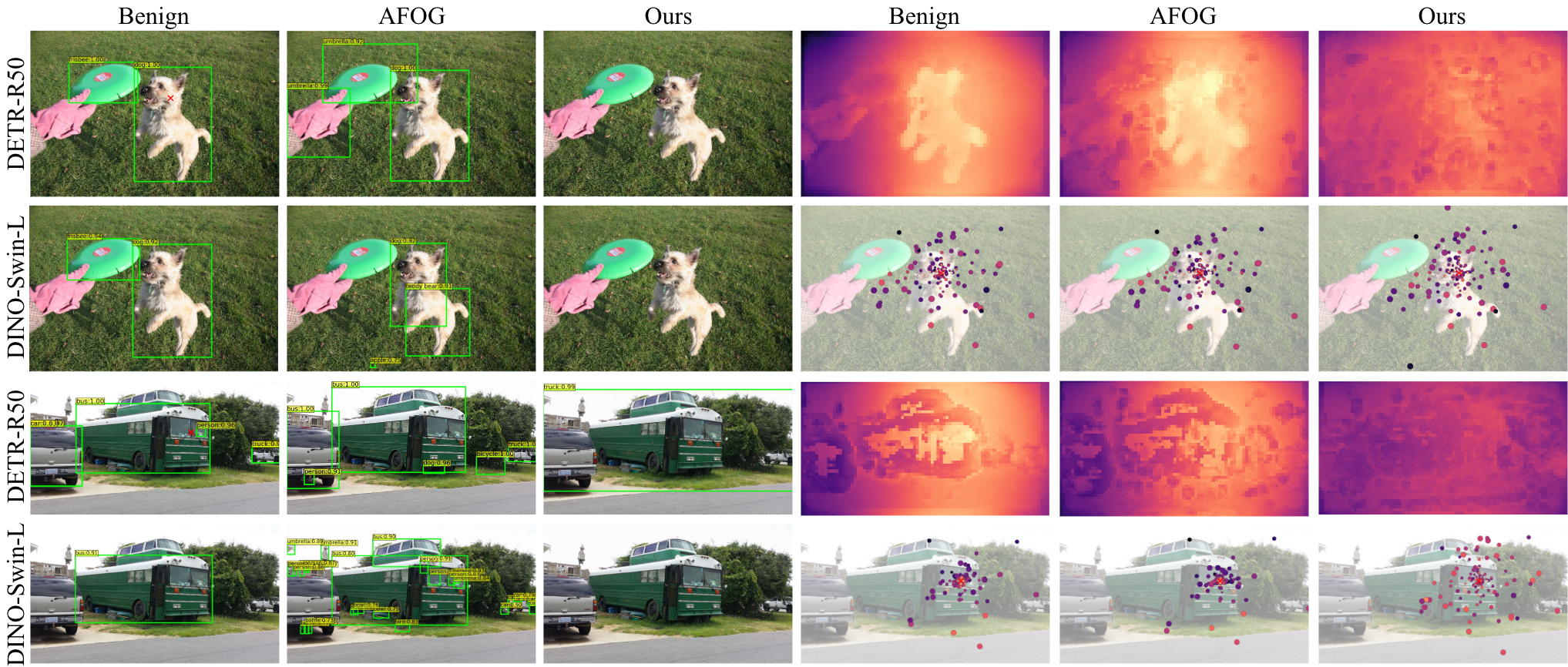}
\caption{Detections (left) and the corresponding final-layer encoder attention (right). For DETR-R50 we show the self-attention averaged over heads and for DINO-Swin-L, the deformable sampling locations coloured by their attention weight.}
\label{fig:qualitative}
\end{figure*}

Surrogate methods leave DETR-R50 largely intact. The strongest of them, GARSDC~\cite{liang2022large}, reaches $6.0$ mAP, but does so with more than $3{,}000$ iterations. The remainder~\cite{li2023improving,chen2020relevance,wang2024gradient,wei2018transferable,xie2017adversarial,li2018robust} sit between $12.7$ and $28.6$ mAP despite budgets twice ours and, in most cases, five times as many iterations. This is consistent with the observation that adversarial examples transfer poorly from convolutional detectors to the DETR family~\cite{nazeri2026evaluating}. Victim model attacks are stronger, yet most remain far from collapsing the detector.  EBAD~\cite{cai2023ensemble}, AttentionFool~\cite{lovisotto2022give}, and OATB~\cite{leng2023object} recover $34.9$, $21.0$, and $26.6$ mAP respectively, OATB with a budget of $0.078$, roughly $2.5\times$ our own. 

AFOG~\cite{yahn2025adversarial} is the only prior method that brings DETR-R50 near collapse, reaching $4.1$ mAP at $\epsilon = 0.031$ and $10$ iterations. Under exactly that budget and schedule, our attack reaches $0.97$ mAP on DETR-R50 with dispersion set as the corruption target, a $\sim4\times$ reduction in mAP relative to AFOG and a $97.7\%$ drop from the clean $42.1$. On DINO-Swin-L the corresponding figures are $1.44$ with re-ranking set as the corruption target, against AFOG's $7.3$. We further conduct a qualitative analysis to visualize both the detection collapse and the attention collapse, as shown in Figure~\ref{fig:qualitative}. AFOG optimizes a detection-loss objective, and attention collapse is only a by-product of that optimization. Their own failure-case analysis~\cite{yahn2025adversarial}, as well as our Figure~\ref{fig:qualitative}, shows that when the AFOG attack fails, the encoder's attention structure is not fully collapsed. In contrast, our attack sets the objective directly on the attention corruption by driving the attention map to collapse hence the detection collapses with it. This highlights that targeting attention is more effective. 

\subsection{Effectiveness Across Detection Transformers}
\label{sec:across_models}

To show that our attack generalizes across different encoder attention mechanisms and different backbones, we run a second set of experiments over a broader family of detection transformers. Table~\ref{tab:transformer_backbones} reports the results for eight models, covering both attention formulations used across the DETR family and a range of backbone architectures. The two DETR models~\cite{carion2020end} use global self-attention and they are marked with (*). The other methods are based on DINO with deformable attention in the encoder and with the listed backbones: R50~\cite{he2016deep}, ConvNeXt~\cite{liu2022convnet}, Swin-L~\cite{liu2021swin}, InternImage~\cite{wang2023internimage}, FocalNet~\cite{yang2022focal}, and EVA~\cite{fang2023eva}. In this experiment we fix dispersion as the corruption target across all models. Our attack achieves State-of-the-art over seven models with significant margins. The benign accuracy grows with model capacity but still the encoder attention becomes attack vulnerable where the accuracy drops by  $81\%$ for EVA, the model with highest mAP which has over one billion parameters.

\begin{table}[t]
\caption{Effect of the attention corruption target on DETR-R50 and DINO-Swin-L.}
\label{tab:detr_distortion_modes}
\centering
\begin{tabular}{lcc}
\toprule
Corruption Target & DETR & DINO \\
\midrule
Clean     & 42.1          & 56.8 \\
Random    & 39.9          & 56.2   \\
Dispersion   & 0.97 & 4.73 \\
Re-ranking    & 1.28          & 1.44   \\
Peak-suppression  & 2.48          & 1.59   \\
Permutation   & 2.58          & 9.22   \\
\bottomrule
\end{tabular}
\end{table}

\subsection{Robustness to the Choice of Corruption}
We test whether the vulnerability depends on how the attention is disrupted and the results are shown in Table~\ref{tab:detr_distortion_modes}. On DETR-R50, all four operators reduce detection from a clean 42.1 to below 3 mAP, each surpassing the prior state-of-the-art, with dispersion being strongest at 0.97. A random perturbation of the same budget leaves detection almost unchanged at 39.9. On DINO-Swin-L the same pattern holds, with three of the four operators surpassing the prior state of the art. Dispersion, re-ranking, and peak-suppression reduce detection from 56.8 to 4.73, 1.44, and 1.59 respectively. Permutation is the weakest at 9.22, the only operator not to beat AFOG, yet it still degrades detection by over 80\% from the clean baseline and far below the random control at 56.2. Across both the dense and deformable settings, every structured target collapses detection while random noise does not, implying that the vulnerability lies in disrupting the attention distribution itself rather than in any single corruption objective.

\begin{table}[t]
\caption{Layer-wise effectiveness of the dispersion attack on DETR-R50. Each encoder layer is targeted individually, then all six jointly.}
\label{tab:detr_layerwise}
\centering
\begin{tabular}{lccccccc}
\toprule
Layer & L1 & L2 & L3 & L4 & L5 & L6 & All \\
\midrule
mAP & 3.60 & 4.61 & 7.36 & 3.36 & 2.57 & \textbf{0.97} & 1.03 \\
\bottomrule
\end{tabular}
\end{table}

\subsection{Ablation Study}
\label{sec:ablation}

\begin{table}[t]
\caption{Cross-model transferability across DETR-R50 and DINO-R50. \emph{source}$\rightarrow$\emph{target} gives the direction.}
\label{tab:transfer}
\centering
\begin{tabular}{llcc}
\toprule
Direction & Attack & Source & Transfer \\
\midrule
\multirow{2}{*}{DETR $\rightarrow$ DINO}
  & AFOG          & 4.1           & 17.81 \\
  & \textbf{Ours} & \textbf{0.97} & \textbf{12.19} \\
\midrule
\multirow{2}{*}{DINO $\rightarrow$ DETR}
  & AFOG          & 5.3           & 13.20 \\
  & \textbf{Ours} & \textbf{4.18} & \textbf{12.35} \\
\bottomrule
\end{tabular}%
\end{table}

We now isolate the factors behind the attack's effectiveness. We study its transferability across models, the encoder layer targeted, the choice of optimizing in pre and post softmax space, and the effect of the perturbation budget and iteration count. Unless otherwise stated we set dispersion as the corruption target on the final encoder layer.

\subsubsection{Attack Transferability Across Detection Transformers}
We test black-box transferability by crafting the perturbation on one model and evaluating it on the other, in both directions between DETR-R50 and DINO-R50 (Table~\ref{tab:transfer}). As expected, transfer is weaker than the white-box setting for both attacks, but our perturbations transfer more effectively than AFOG in both directions, with the larger margin when transferring from DETR's dense attention to DINO's deformable attention.

\subsubsection{Encoder Layer-wise Effectiveness}
Table~\ref{tab:detr_layerwise} targets each encoder layer individually and all six jointly. The final layer is by far the strongest single target at 0.97 mAP, as it feeds the decoder directly. Attacking all six layers jointly is competitive but no stronger, so targeting the final layer alone is the more efficient choice.

\begin{table}[t]
\caption{Effect of optimizing the corruption objective in pre-softmax logit space versus post-softmax attention space on DETR-R50.}
\label{tab:pre_post_softmax}
\centering
\begin{tabular}{lcc}
\toprule
Corruption Target & Pre-softmax & Post-softmax \\
\midrule
Dispersion  & \textbf{0.97} & 7.96 \\
Re-ranking  & \textbf{1.28} & 7.45 \\
\bottomrule
\end{tabular}
\end{table}

\begin{figure}[t]
  \centering
  \includegraphics[width=\columnwidth]{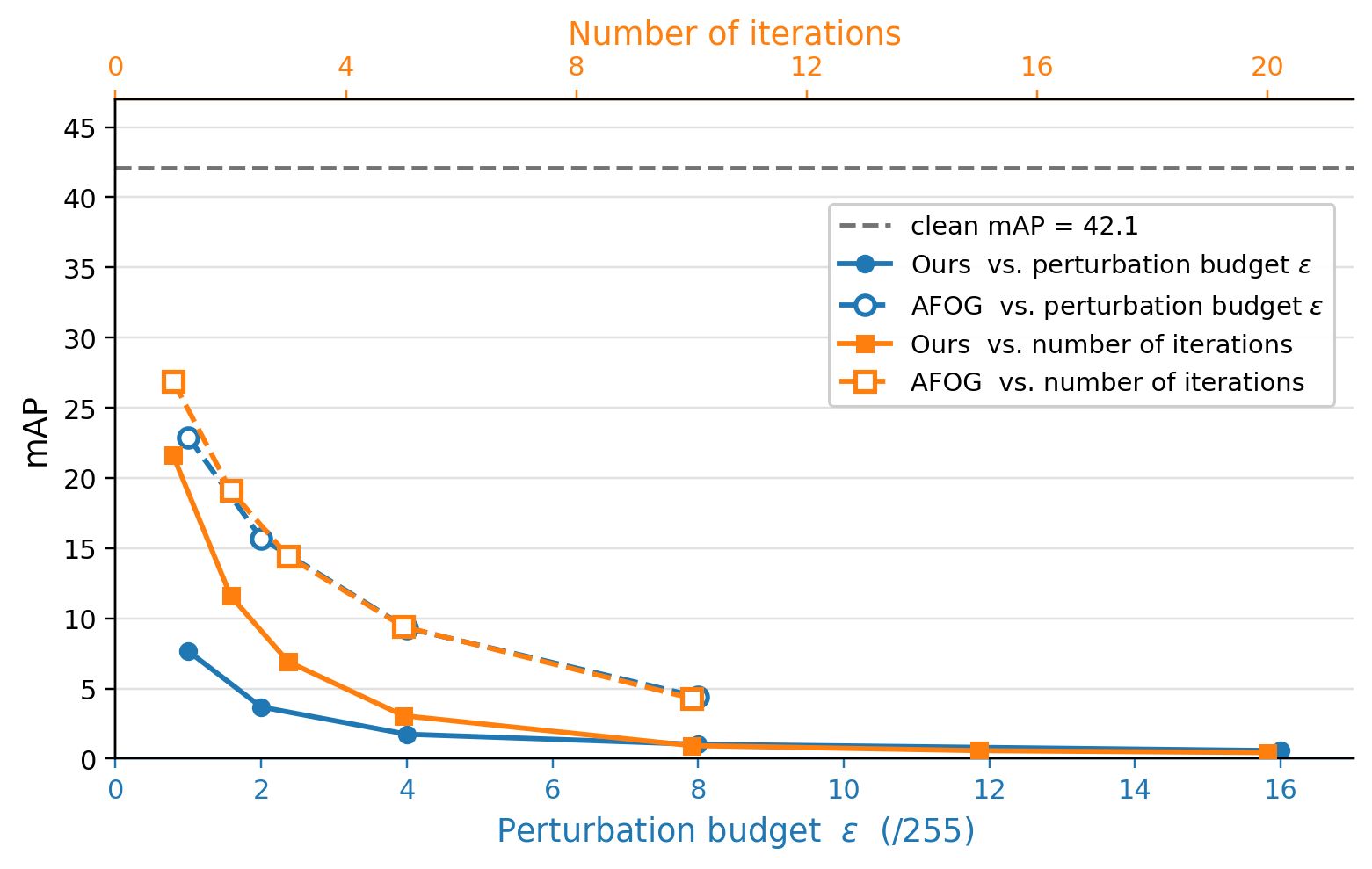}
 \caption{Adversarial mAP on DETR-R50 versus perturbation budget and iteration count , each swept with the other at its default. Ours (solid) dominates AFOG (dashed) along both axes.}
  \label{fig:ablation_sweeps}
\end{figure}

\subsubsection{Optimization Space}
As described in Methodology, we optimize on the pre-softmax logits rather than the post-softmax attention map, and we test this empirically. Table~\ref{tab:pre_post_softmax} shows the pre-softmax objective is stronger and optimizing the same targets in probability space is six to eight times weaker.

\subsubsection{Perturbation Budget and Iteration Count}
Figure~\ref{fig:ablation_sweeps} sweeps the two axes of attack cost, the perturbation budget $\epsilon$ and the iteration count. Our attack dominates AFOG across both sweeps. Even at a quarter of the budget or half the iterations, our attack surpasses AFOG, the previous state of the art, run at its full budget and full iteration count, demonstrating the effectiveness of our attack.

\subsubsection{Attention Collapse Over Attack Iterations}
\begin{figure}[t]
\centering
\includegraphics[width=\columnwidth]{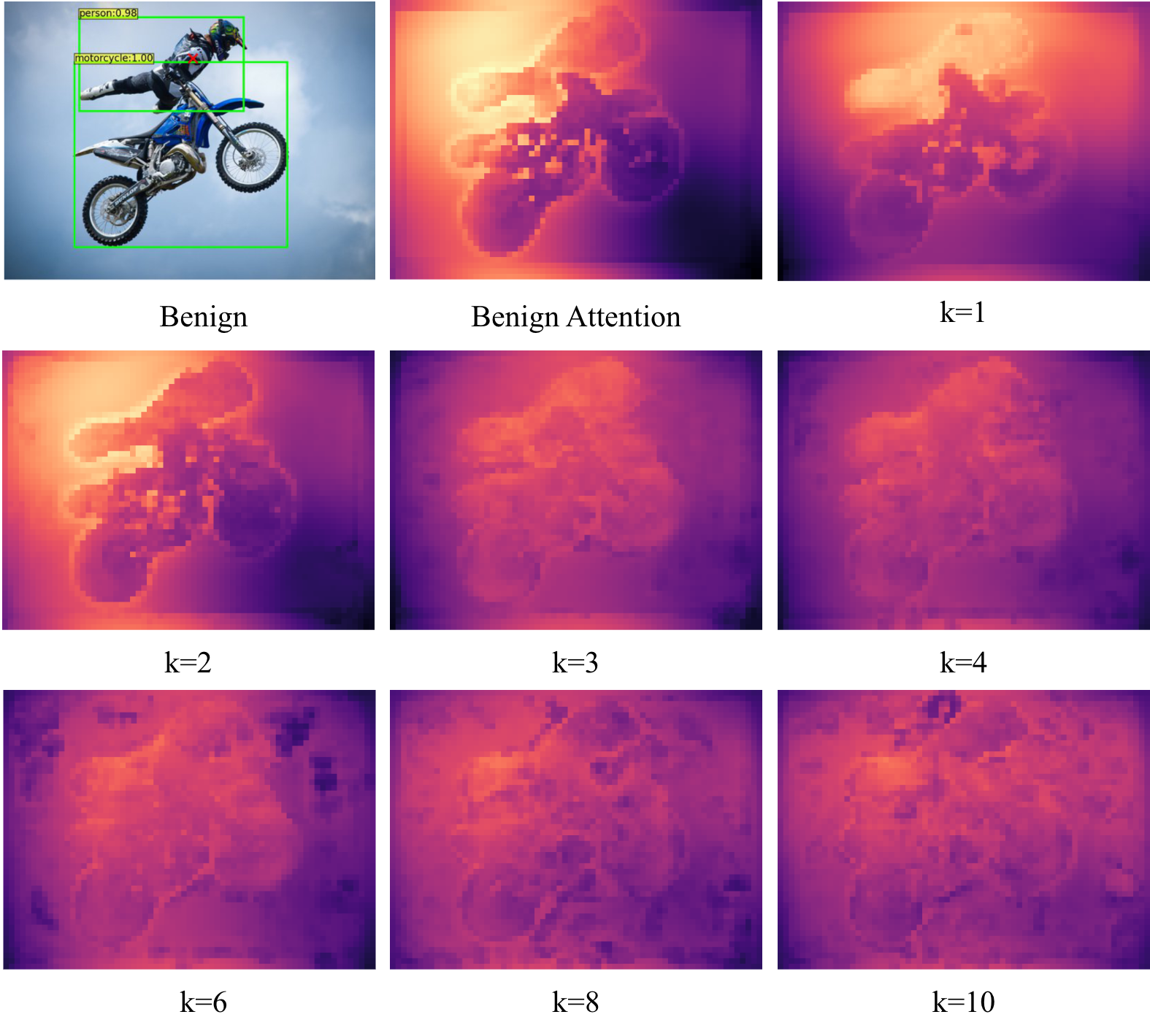}
\caption{Last-layer encoder attention on DETR-R50 over optimization iterations $k$ with perturbation budget fixed at $\epsilon = 8/255$. The attention degrades progressively and collapses to a structureless map within a few iterations.}
\label{fig:supp_iterations}
\end{figure}

The iteration sweep in Figure~\ref{fig:ablation_sweeps} shows that adversarial mAP falls sharply within the first few optimization steps, surpassing the previous state of the art at half the iterations. Figure~\ref{fig:supp_iterations} visualizes what happens to the encoder attention over these iterations. As the attack progresses the attention degrades steadily, and within a few iterations it is flattened across the image. This shows that targeting the attention corrupts it efficiently, collapsing detection within only a few iterations.

\section{Conclusion}
We introduced an evasion attack that corrupts a detection transformer's encoder attention directly, using an imperceptible perturbation that drives the attention toward a structured target with no patch and no reliance on the detection loss. Under an identical budget and iteration count, the attack reduces DETR-R50 to 0.97 mAP, roughly four times lower than the strongest prior attack. Our attack also degrades detection across eight detectors with different backbones, spanning both dense and deformable attention where it reduces DINO-Swin-L from 56.8 mAP to 1.44. The four qualitatively distinct corruption targets are all effective, indicating that the vulnerability lies in disrupting the attention rather than in any single objective. To our knowledge, this is the first work to expose the encoder attention of detection transformers as a direct and imperceptible attack surface, and to show empirically how readily it can be corrupted. In future work, it would be interesting to investigate the fundamental reasons behind this vulnerability, to characterize how corrupted attention propagates to the detector's predictions, and to design defenses that make encoder attention robust.

\bibliographystyle{IEEEtran}
\bibliography{ref}

\end{document}